\def\year{2020}\relax
\documentclass[letterpaper]{article} 
\usepackage{aaai20}  
\usepackage{times}  
\usepackage{helvet} 
\usepackage{courier}  
\usepackage[hyphens]{url}  
\usepackage{graphicx} 
\usepackage{graphicx}  
\usepackage{amssymb}
\usepackage{booktabs}
\usepackage{amsfonts}
\usepackage{amsmath}
\usepackage{algorithm}
\usepackage{algpseudocode}

\title{Instance Enhancement Batch Normalization: \\ an Adaptive Regulator of Batch Noise}

\author{Senwei Liang\textsuperscript{\rm 1}\thanks{Equal contribution},
Zhongzhan Huang\textsuperscript{\rm 2$*$}, Mingfu Liang\textsuperscript{\rm 3},
Haizhao Yang\textsuperscript{\rm 1,4}\\
\textsuperscript{\rm 1}{National University of Singapore} \\
\textsuperscript{\rm 2}{New Oriental AI Research Academy}\\
\textsuperscript{\rm 3}{Northwestern University}\\
\textsuperscript{\rm 4}{Purdue University}\\
liangsenwei@u.nus.edu,\quad
hzz\_dedekinds@foxmail.com,\\
mingfuliang2020@u.northwestern.edu,\quad
matyh@nus.edu.sg}

\begin{document}

\maketitle

\begin{abstract}

Batch Normalization~(BN)~\cite{Ioffe:2015:BNA:3045118.3045167} normalizes the features of an input image via statistics of a batch of images and hence BN will bring the noise to the gradient of the training loss.
Previous works indicate that the noise is important for the optimization and generalization of deep neural networks, but too much noise will harm the performance of networks. In our paper, we offer a new point of view that self-attention mechanism can help to regulate the noise by enhancing instance-specific information to obtain a better regularization effect.
Therefore, we propose an attention-based BN called Instance Enhancement Batch Normalization~(IEBN) that recalibrates the information of each channel by a simple linear transformation.
IEBN has a good capacity of regulating noise and stabilizing network training to improve generalization even in the presence of two kinds of noise attacks during training. Finally, IEBN outperforms BN with only a light parameter increment in image classification tasks for different network structures and benchmark datasets.

\end{abstract}

\section{Introduction}
Mini-batch Stochastic Gradient Descent~(SGD) is an effective method in large-scale optimization by aggregating multiple samples at each iteration to reduce operation and memory cost. However, SGD is sensitive to the choice of hyperparameters and it may cause training instability~\cite{luo2018adaptive}. Normalization is one possible choice to remedy SGD methods for better stability and generalization. Batch Normalization~(BN)~\cite{Ioffe:2015:BNA:3045118.3045167} is a frequently-used normalization method that normalizes the features of an image using the mean and variance of the features of a batch of images during training. Meanwhile, the tracked mean and variance that estimate the statistics of the whole dataset are used for normalization during testing. It has been shown that BN is an effective module to regularize parameters~\cite{luo2018towards}, stabilize training, smooth gradients~\cite{shibani2018how}, and enable a larger learning rate ~\cite{nils2018understanding,cai2019bn} for faster convergence. Further, the generation of noise and the usage of both SGD and BN are inseparable.

Two kinds of noise effected in SGD and BN are concerned in this paper.

\noindent\textbf{Estimation Noise.} In BN, the mean and variance of a batch are used to estimate those of the whole dataset; in SGD, the gradient of the loss over the batch is applied to approximate that of the whole dataset. These estimation errors are called estimation noise.

\noindent\textbf{Batch Noise.} In the forward pass, the features of an instance are incorporated with batch information via the normalization with batch statistics. The gradient of the loss of an instance is disturbed by the batch information due to the forward pass. These disturbances to an instance caused by the batch is referred to as batch noise.

The randomness of BN and SGD has been well-known to improve the performance of deep networks. BN with the estimation noise can work as an adaptive regularizer of parameters~\cite{luo2018towards} and the moderate noise can help escape bad local minima and saddle point~\cite{jin2017escape,ge2015escaping}. There exists extensive study on optimizing effectiveness via tuning batch sizes. On the one hand, a small batch size will lead to high variance of statistics and weaken the training stability~\cite{we2018gn}. On the other hand, a large batch size can reduce the estimation noise but it will cause a sharp landscape of loss~\cite{Keskar2016On} making the optimization problem more challenging.
Therefore, an appropriate batch size is required to make a good balance. At that time, the estimation noise becomes moderate when batch size is not small ~\cite{luo2018towards} and the batch noise turns into the primary noise.

The appropriate batch noise is necessary for a model. In fact, too much noise may harm the performance of the network. In our paper, we design two kinds of additional noise injected into the network to heighten the batch noise. Table~\ref{tab:Constant noise} and Table~\ref{tab:tri_attack} show the performance of the networks with BN under the noise attack. And we can see the networks with BN can not resist augmented batch noise, which results in the instability of training and degradation of performance.

\begin{figure}
  \centering
  \includegraphics[height=2in, width=3in]{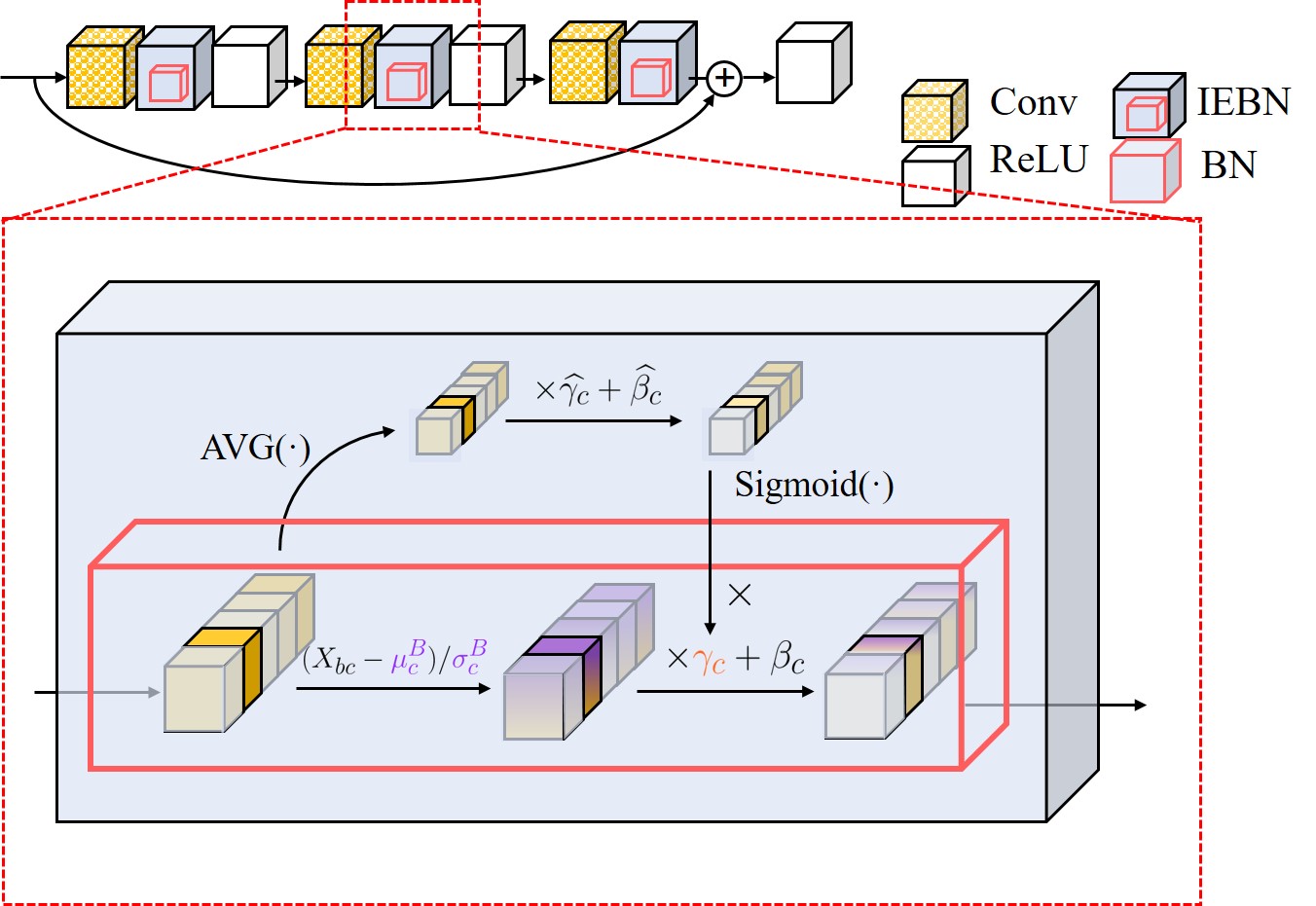}\\
  \caption{The illustration of IEBN. The top shows a block of ResNet. The bottom is the showcase of IEBN, where the box with red border is the basic flow of BN. $\text{AVG}(\cdot)$ means the average pooling over a channel.}
  \label{fig:iebn_structure}
\end{figure}

It remains a problem on how to infuse a model with the appropriate batch noise. Through experiments on style transfer, we find that self-attention mechanism is an adaptive batch noise regulator for the model by enhancing instance specificity, which motivates us to design a new normalization to combine the advantage of BN and self-attention. This paper proposes an attention-based BN which adaptively emphasizes instance information called as Instance Enhancement Batch Normalization~(IEBN). The idea behind IEBN is simple. As shown in Fig.~\ref{fig:iebn_structure}, IEBN extracts the instance statistic of a channel before BN and applies it to rescale the output channel of BN with a pair of additional parameters for each channel. IEBN costs a light parameter increment and it can be plugged into existing architectures easily. The extended experiment shows that IEBN outperforms BN on benchmark datasets over popular architectures for image classification. Our contribution is summarized as followed,

\begin{enumerate}
    \item We offer a point of view that self-attention mechanism can regulate the batch noise adaptively.
    \item We propose a simple-yet-effective and attention-based BN called as Instance Enhancement Batch Normalization~(IEBN). We demonstrate the effectiveness of it on benchmark datasets with different network architectures.
\end{enumerate}

    \begin{algorithm}[ht]
        \caption{Instance Enhancement Batch Normalization}
        \label{alg:IEBN}
        \begin{algorithmic}[1]
            \Require
            $X$ is a batch input of size $B \times C \times H \times W$;

            Paramentes: $\gamma_c$, $\beta_c$, $\hat{\gamma}_c$ and $\hat{\beta}_c$, $c=1,\cdots,C$;

            \Ensure
            $\{$$Y$ = $\textbf{IEBN}_{\gamma_{c},\beta_{c},\hat{\gamma}_{c}, \hat{\beta}_{c}}(X)$$\}$;
            \State $\hat{\gamma}_c \gets 0$;$\hat{\beta}_c \gets -1$
            \For{channel $c$ from 1 to C}
            \State$\mu_c^B \gets \frac{1}{B\cdot H\cdot W}\sum\limits_{b=1}^B\sum\limits_{h=1}^H\sum\limits_{w=1}^W X_{bchw}$ \label{alg:step:mean}
            \State $\sigma_c^B \gets \sqrt{\frac{1}{B\cdot H\cdot W}\sum\limits_{b=1}^B\sum\limits_{h=1}^H\sum\limits_{w=1}^W (X_{bchw}-\mu_c^B)^2+\epsilon}$\label{alg:step:std}
                \For{instance $b$ from 1 to B}
                    \State $\delta_{bc}$ $\gets$  $Sigmoid(\text{AVG}(X_{bc})\times \hat{\gamma}_c + \hat{\beta}_c)$ \label{lst:line:test}
                    \State $\hat{X}_{bc}$ $\gets$  $ (X_{bc}-\mu_c^B)/\sigma_c^B$
                    \State $Y_{bc} \gets \hat{X}_{bc} \times (\gamma_c \times \delta_{bc})+\beta_c$

                \EndFor
            \EndFor
        \end{algorithmic}
    \end{algorithm}

\section{Related Work}
This session reviews related works in two fields: normalization and self-attention mechanism, especially those combining these techniques.

\noindent\textbf{Normalization.} The normalization layer is an important component of a deep network. Multiple normalization methods have been proposed for different tasks. Batch Normalization~\cite{Ioffe:2015:BNA:3045118.3045167} which normalizes input by mini-batch statistics has been a foundation of visual recognition tasks~\cite{he2016deep}. Instance Normalization~\cite{ulyanov2017in} performs one instance BN-like normalization and is widely used in generative model~\cite{justin2016faster,zhu2017cyclegan}. There are some variants of BN, such as Conditional Batch Normalization~\cite{de2017cbn} for Visual Questioning and Answering, Group Normalization~\cite{we2018gn} and Batch Renormalization~\cite{Ioffe2017BRN} for small batch size training. Also there are Adaptive Batch Normalization~\cite{li2018adaptive} for domain adaptation and Switchable normalization~\cite{luo2018differentiable} which learns to select different normalizers for different normalization layers. Among them, Conditional Batch Norm and Batch Renorm adjust the trainable parameters in reparameterization step of BN, which is similar to our work that modifies the trainable scaling parameter.

\noindent\textbf{Self-attention Mechanism.} Self-attention mechanism selectively focuses on the most informative components of a network via self-information processing and has gained a promising performance on vision tasks. The procedure can be divided into three parts. First, the added-in module extracts internal information of a networks which can be squeezed channel-wise information~\cite{hu2018squeeze,li2019selective} or spatial information~\cite{wang2018non,li2019sge}. Next, the module processes the extraction and generates a mask to measure the importance of features via fully connected layer~\cite{hu2018squeeze}, convolution layer~\cite{wang2018non} or LSTM~\cite{huang2019dia}. Last, the mask is applied back to features to adjust feature importance.

\noindent\textbf{Comparison with recent work.} The cooperation of BN and attention dates back to Visual Questioning and Answering (VQA). For this task, Conditional Batch Norm~\cite{de2017cbn} is proposed to influence the feature extraction of an image via the feature collected from the question.
Note that for VQA, the features from question can be viewed as external attention to guide the training of overall network. In our work, the IEBN we proposed can also be viewed as a kind of Conditional Batch Norm but the guidance of the network training is using the internal attention since we extract the information from the image itself.

There is a recent work that combines normalization and self-attention together~\cite{Jia_2019_CVPR}. Different from previous works that generate the rescaling coefficient of normalization by modeling the relationship between channels, we generated a rescaling coefficient independently for each channel.

\section{BN and SE modules}
This session first reviews BN and then reviews Squeeze-and-Excitation~(SE) module~\cite{hu2018squeeze} which is regarded as the representative of self-attention mechanism and is used for the analysis of self-attention in our paper.

We consider a batch input $X \in {\mathbb{R}^{B \times C \times H \times W}}$, where $B, C, H$ and $W$ stand for batch size, number of channels (feature maps), height and width respectively. For simplicity, we denote $X_{bchw}=X[b,c,h,w]$ as the value of pixel $(h,w)$ at the channel $c$ of the instance $b$ and $X_{bc} = X[b,c,:,:]$ as the tensor at the channel $c$ of the instance $b$.

\subsection{Review of BN and SE module}
\noindent\textbf{BN.} The computation of BN can be divided into two steps: batch-normalized step and reparameterization step. Without loss of generality, we perform BN on the channel $c$ of the instance $b$, i.e., $X_{bc}$. In batch-normalized step, each channel of features are normalized using mean and variance of a batch,
\begin{align}
    \hat{X}_{bc} = (X_{bc}-\mu_c^{B})/\sigma_c^B,
    \label{eqn:batch_norm}
\end{align}
where $\mu_c^{B}, \sigma_c^B$ are defined in Step~\ref{alg:step:mean} and Step~\ref{alg:step:std} in Alg.\ref{alg:IEBN} as the estimation of mean and standard derivation respectively of the whole dataset. Then in reparameterization step, a pair of learnable parameters $\gamma_c, \beta_c$ scale and shift the normalized tensor $\hat{X}_{bc}$ to restore the representation power,
\begin{align}
    \hat{Y}_{bc} = \hat{X}_{bc}\times \gamma_c + \beta_c.
    \label{eqn:reparams}
\end{align}
As said in Introduction, the batch noise mainly comes from the batch-normalized step where the feature of the instance $b$ is mixed with information from the batch, i.e., $\mu_c^{B}$ and $\sigma_c^B$.
\noindent\textbf{SE module.} SE module utilizes the relationship between all channels in the same layer and generates a reweighing coefficient for each channel. Consider that SE module is connected to $X$ and recalibrates channel information of $X$.

First, the SE module squeezes the information of channels for the instance $b$ by taking the average over each channel,
\begin{align}
  m_{bc} = \text{AVG}(X_{bc}) = \frac{1}{H\cdot W}\sum_{h=1}^H\sum_{w=1}^W X_{bchw},
  \label{eqn:m}
\end{align}
where $c=1,\cdots,C$ goes through all channels. Then, a fully connected layer stacking of a linear transformation, ReLU activation and a linear transformation with learnable parameters $\mathbf{W}$ is used to fuse the information of channels. Then Sigmoid function (i.e., $\text{sig}(z) = 1/(1+e^{-z})$) is applied to the value and we get the rescaling coefficient for each channel. We denote the fully connected layer as $\text{FC}(\cdot;\mathbf{W})$ which maps $\mathbb{R}^C$ to $\mathbb{R}^C$. The rescaling coefficient is
\begin{align}
    [\hat{\delta}_{b1}; \cdots; \hat{\delta}_{bC}] = \text{sig}(\text{FC}([m_{b1}; \cdots; m_{bC}];\mathbf{W})).
    \label{eqn:sedelta}
\end{align}
The rescaling coefficient is applied to its corresponding channel to adjust the the channel, and we obtain $X_{bc} \times \hat{\delta}_{bc}$.
\subsection{Adaptive batch noise regulation via SE module}
We explore the role of self-attention mechanism on regulating batch noise through the the style transfer task~\cite{gatys2016image}. We use the style transfer method which generates an image by a network called transformation network~\cite{johnson2016perceptual}.

\begin{figure*}
  \centering
  \includegraphics[height=5in, width=6in]{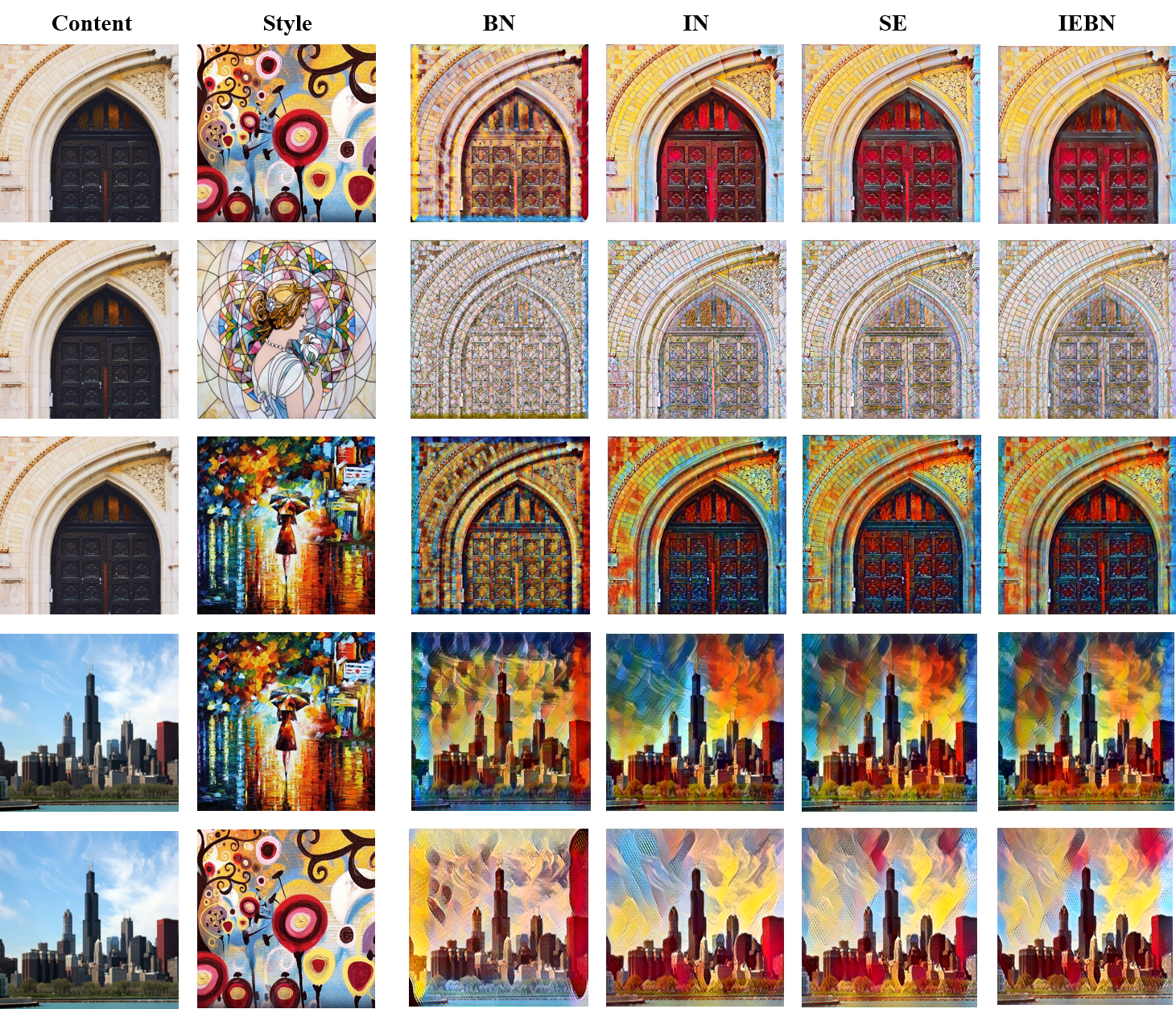}\\
  \caption{Stylization results obtained by applying style (second column) to  content images (first column) with different normalization methods. Specially, ``SE'' means the transformation network with BN and SE module. The style of the generated images with BN appears more confused, but those with SE or IEBN are quite similar to IN visually.}\label{fig:styletrans1}
\end{figure*}

It has been empirically shown that the type of normalization in the network has an impact on the quality of image generation~\cite{ulyanov2017in,huang2017arbitrary,dumoulin2016learned}. Instance Normalization~(IN) is widely used in generative models and it had proved to have a significant advantage over BN in style transfer tasks~\cite{ulyanov2017in}. The formulation of IN is followed,
\begin{equation}
        \left(\frac{X_{bc}-\mu(X_{bc})}{\sigma(X_{bc})}\right)\cdot \gamma + \beta=\frac{\gamma}{\sigma(X_{bc})}\cdot X_{bc} + \beta-\frac{\mu(X_{bc})}{\sigma(X_{bc})}\cdot \gamma,
        \label{eqn:instance norm}
\end{equation}
where $\mu(X_{bc})$ and $\sigma(X_{bc})$ denote the mean and standard deviation of the instance $b$ at the channel $c$. Similarly, the formulation of BN can be written in this form,
\begin{equation}
        \left(\frac{X_{bc}-\mu_c^{B}}{\sigma_c^B}\right)\cdot \gamma + \beta=\frac{\gamma}{\sigma_c^B}\cdot X_{bc} + \beta-\frac{\mu_c^{B}}{\sigma_c^B}\cdot \gamma.
        \label{eqn:batch norm}
\end{equation}
$\gamma$ and $\beta$ are learned parameters and both are closely related to the target style~\cite{dumoulin2016learned}. From Eqn.~\ref{eqn:instance norm} and Eqn.~\ref{eqn:batch norm}, IN or BN directly leads to the scaling of $\gamma$ that affects the style of images. Different from BN, IN affects the style by self-information instead of batch information. Fig.~\ref{fig:styletrans1} compares the quality of images generated by the network with BN and IN. Note that, the style transfer task is noise-sensitive, which coincides with \cite{salimans2016weight}. When the batch noise is added by BN, the style of the generated images becomes more confused.

\begin{figure*}
        \centering
        \includegraphics[height=2in, width=7in]{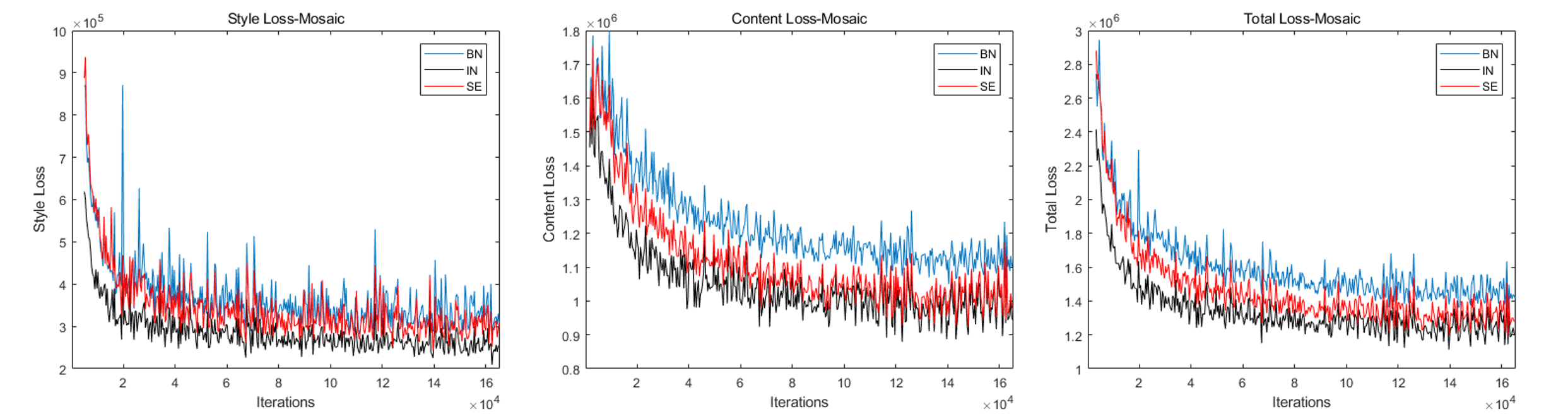}
        \caption{Training curves of style transfer networks with Mosaic style and different normalization methods. Specially, ``SE'' means the transformation network with BN and SE module. }
        \label{fig:styletrans2}
    \end{figure*}

We add the SE module to the transformation network with BN to find its effectiveness of regulating batch noise. In Fig.~\ref{fig:styletrans1}, the attention mechanism~(SE) visually improves the effect of style transfer and the quality of the generated images is closer to that of IN.  Fig.~\ref{fig:styletrans2} shows the training loss against the number of iterations by applying the style Mosaic. The BN network with SE modules achieves smaller style loss and smaller content loss than the original BN network and is closer to an IN network~(see Appendix for more results about the loss by applying other styles). For a brief analysis of the effects of self-attention mechanism, we consider such structure, i.e., the SE module is plugged into the residual part of the residual block where the last module is BN. The formulation of plugging SE module into BN network is as followed,
\begin{equation}
        \left(\frac{X_{bc}-\mu_c^B}{\sigma_c^B}\right)\cdot \gamma \hat{\delta}_{bc} + \beta\hat{\delta}_{bc}=\frac{\gamma\hat{\delta}_{bc}}{\sigma_c^B}\cdot X_{bc} + \beta\hat{\delta}_{bc}-\frac{\mu_c^B}{\sigma_c^B}\cdot \gamma\cdot \hat{\delta}_{bc},
        \label{eqn:se-style}
\end{equation}
where $\hat{\delta}_{bc}$ is rescaling coefficient generated by applying SE module to $\hat{Y}_b$ in Eqn.~\ref{eqn:reparams}. Although BN introduces batch noise to an instance, we can see that $\hat{\delta}_{bc}$ helps adjust the batch statistics after applying SE module to BN in Eqn.~\ref{eqn:se-style}. The attention mechanism such as SE module may be good at alleviating the batch noise and we will investigate it further.

\begin{table*}[htbp]
  \centering
    \begin{tabular}{|l|c|c|c|c|c|c|c|}
    \toprule
          & Dataset & \multicolumn{2}{c|}{BN} & \multicolumn{2}{c|}{SE} & \multicolumn{2}{c|}{IEBN} \\
\cmidrule{3-8}          &       & \#P(M) & top1-acc. & \#P(M) & top1-acc. & \#P(M) & top1-acc. \\
    \midrule
    ResNet164 & CIFAR100 & 1.73  & 74.29  & 1.93  & 75.80     & 1.75  & 77.09  \\
    PreResNet164 & CIFAR100 & 1.73  & 76.56  & 1.92  & 77.41     & 1.75     &77.27\\
    DenseNet100-12 & CIFAR100 &   0.80 & 77.23 &   0.81    &  78.43     & 0.82       & 78.57 \\
    ResNext29,8x64 & CIFAR100 & 34.52  & 81.47  & 35.04  & 83.05     & 34.57     & 82.45  \\
    \midrule
    ResNet164 & CIFAR10 & 1.70    & 93.93  & 1.91     & 94.65     & 1.73     & 95.03  \\
    Preresnet164 & CIFAR10 & 1.70     & 95.01  & 1.90     & 95.18     & 1.73     & 95.09  \\
    DenseNet100-12 & CIFAR10 &  0.77     & 95.29 &    0.78   &   95.76    &  0.79     & 95.83 \\
    ResNext29,8x64 & CIFAR10 & 34.43     & 96.11  & 34.94     & 96.30    & 34.48     & 96.26 \\
    \midrule
    ResNet34 & ImageNet & 21.81     & 73.91  & 21.97     & 74.39  & 21.82     & 74.38 \\
    ResNet50 & ImageNet & 25.58     & 76.01  & 28.09     & 76.61  & 25.63     & 77.10 \\
    ResNet152 & ImageNet & 60.27     & 77.58  & 66.82     & 78.36  & 60.41     & 79.17 \\
    ResNext50 & ImageNet & 25.03     & 77.19     & 27.56     & 78.04     & 25.09     & 77.99 \\
    \bottomrule
    \end{tabular}%
    \caption{Accuracy (\%) on benchmark datasets with different architectures using BN, SE module or IEBN.}
  \label{tab:experiment}%
\end{table*}%
\section{Instance Enhancement Batch Normalization}
Through experiments on style transfer shown in Fig.~\ref{fig:styletrans1}, we find that self-attention mechanism is an adaptive batch noise regulator for the model by enhancing instance specificity, which motivates us to design a new normalization to combine the advantage of BN and self-attention. This session introduces IEBN and analyzes its role in enhancing instance information in style transfer task. Last, we evaluate the performance of IEBN empirically on image classification task.
\subsection{Formulation of IEBN}
The showcase of IEBN is shown in Fig.~\ref{fig:iebn_structure}, where we highlight the instance enhancement process of one channel. The detailed computation can be found in Alg.~\ref{alg:IEBN}. IEBN is based on the adjustment of the trainable scaling parameter on BN and its implementation consists of three operations: global squeezing, feature processing, and instance embedding.

\noindent\textbf{Global Squeezing.} The global reception field of a feature map is captured by average pooling $\text{AVG}(\cdot)$. We obtain a shrinking feature descriptor $m_{bc}$ of the channel $c$ for the instance $b$, which is the same as Eqn.~\ref{eqn:m}.

\noindent\textbf{Feature Processing.}
Compared with Eqn.~\ref{eqn:sedelta}, we introduce an addition pair of parameters $\hat{\beta}_c$, $\hat{\gamma}_c$ for the $c_{th}$ channel, which serve as scale and shift respectively to linearly transform $m_{bc}$. Then Sigmoid function ($\text{sig}(\cdot)$) is applied to the value after linear transformation as a gating mechanism:
\begin{align}
    \delta_{bc} = \text{sig}(\hat{\gamma}_c m_{bc} + \hat{\beta}_c).
    \label{eqn:delta}
\end{align}
Specially, the parameters $\hat{\gamma}_c,\hat{\beta}_c$ are initialized by constant 0 and -1 respectively. We will discuss it in Ablation Study.

\noindent\textbf{Instance Embedding.} $\delta_{bc}$ works as a weight coefficient to adjust the scaling in the reparameterization step of BN for the instance $b$. We embed the recalibration $\delta_{bc}$ to compensate the instance information in Eqn.~\ref{eqn:reparams},
\begin{align}
Y_{bc} = \hat{X}_{bc} \times (\gamma_c \times \delta_{bc})+\beta_c.
\label{eqn:instance_enhancement}
\end{align}
The $\delta_{bc}$ is composed of nonlinear activation function and an  additional pair of parameters£¬ which helps improve the nonlinearity of reparameterization of BN.

We conduct IEBN on all channels, i.e., $c=1,2,\cdots,C.$ Compared with BN, the parameter increment comes from the additional pair parameter for generating coefficient for each channel. The total number of parameter increment is equal to twice the number of channels.
\subsection{Instance enhancement in style transfer}
IEBN is a BN equipped with self-attention and Fig.~\ref{fig:styletrans1} shows the similarity of the generated images of the SE module and IEBN. In fact, we consider IEBN:
\begin{equation}
        \left(\frac{X_{bc}-\mu_c^B}{\sigma_c^B}\right)\cdot \gamma \delta_{bc} + \beta=\frac{\gamma\delta_{bc}}{\sigma_c^B}\cdot X_{bc} + \beta-\frac{\mu_c^B}{\sigma_c^B}\cdot \gamma\cdot \delta_{bc},
        \label{eqn:iebn-style}
\end{equation}
where $\delta_{bc}$ is defined in Eqn.~\ref{eqn:delta} and $\delta_{bc}$ contains information from the instance $b$. It seems like the added-in $\delta_{bc}$ is only directly applied to scaling parameter $\gamma$ of BN, but it does scale the batch information (i.e., $\mu_c^B, \sigma_c^B$) to regulate the batch information via supplement of instance information. This adjustment of batch information via $\delta_{bc}$ makes the Eqn.~\ref{eqn:iebn-style} closer to Eqn.~\ref{eqn:instance norm} than Eqn.~\ref{eqn:batch norm} and also leads to the similar results in style transfer between IN and IEBN.

\subsection{Experiments}
In this section, we evaluate the effect of IEBN in image classification task and empirically demonstrate its effectiveness.

\noindent\textbf{Dataset and Model.} We conduct experiments on CIFAR10, CIFAR100~\cite{cifar}, and ImageNet 2012~\cite{ILSVRC15}. CIAFR10 or CIFAR100 has 50k train images and 10k test images of size 32 by 32 but has 10 and 100 classes respectively. ImageNet 2012~\cite{ILSVRC15} comprises 1.28 million training and 50k validation images from 1000 classes, and the random cropping of size 224 by 224 is used in our experiments. We evaluates our methods with popular networks, ResNet~\cite{he2016deep}, PreResNet~\cite{he2016deep} and ResNeXt~\cite{xie2017aggregated}. In our experiments, we replace all the BNs in the original networks with IEBN. The implementation details can be found in the Appendix.

\noindent\textbf{Image Classification.} As shown in Table \ref{tab:experiment}, the IEBN improves the testing accuracy over BN for different datasets and different network backbones. For small-classes dataset CIFAR10, the performance of the networks with BN is good enough, so there is not large space for improvement. However, for CIFAR100 and ImageNet datasets, the networks with IEBN achieve a significant testing accuracy improvement over BN. In particular, the performance improvement of the ResNet with the IEBN is most remarkable.
\section{Analysis}
In this session, we explore the role of self-attention mechanism on enhancing instance information and regulating the batch noise. We analysis through experiments with two kinds of noise attack designed in our paper.

\subsection{Constant Noise Attack} We add constant noise into every BN of the network in batch-normalized step as followed, 
\begin{align}
    \hat{X}_{bc} = [(X_{bc}-\mu_c^{B})/\sigma_c^B ]\cdot N_a + N_b,
    \label{eqn:Constant noise}
\end{align}
where $(N_a, N_b)$ are a pair of constant as the constant noise. Table~\ref{tab:Constant noise} shows the testing accuracy of ResNet164 on CIFAR100 under different pairs of constant noise.

The added constant noise is equivalent to disturbing $\mu_c^{B}$ and $\sigma_c^B$ such that we can use the inaccurate estimations of mean and variance respectively of the whole dataset in training. Denote $(X_{bc}-\mu_c^{B})/\sigma_c^B$ as $\Delta$. Then in the reparameterization step of BN, we introduce the learnable parameters $\gamma$ and $\beta$ and get
\begin{align}
\begin{split}
    \hat{X}_{bc} &= (\Delta \cdot N_a + N_b)\cdot \gamma + \beta\\
              &= \Delta \cdot (N_a \cdot \gamma) + (N_b\cdot \gamma + \beta)
    \label{eqn:Constant noise2}
    \end{split}
\end{align}
 From the inference of Eqn.~\ref{eqn:Constant noise2}, the impact of constant noise can be easily neutralized by the linear transformation of $\gamma$ and $\beta$ because $N_a$ and $N_b$ are just constants. However, in Table~\ref{tab:Constant noise}, the network with only BN is not good at handling most constant noise ($N_a,N_b$). The trainable $\gamma$ and $\beta$ of BN does not have enough power to help BN reduce the impact of the constant noise. Due to the forward propagation, the noise will accumulate as the depth increases and a certain amount of noise leads to poor performance and training instability. As shown in Table~\ref{tab:Constant noise}, SE module can partly alleviate this problem, but not enough as we can see the high variance of the testing accuracy under most pairs of constant noise.
 \begin{table}[htbp]
  \centering
    \begin{tabular}{|c|c|c|c|}
    \toprule
    $(N_a, N_b)$ & BN    & SE    & IEBN \\
    \midrule
    (0.0,0.0) & 74.29$_{(\pm0.64)}$     & 75.80$_{(\pm0.25)}$      & 77.09$_{(\pm0.15)}$ \\
    (0.8,0.8) & 45.42$_{(\pm31.42)}$     & 73.18$_{(\pm0.66)}$      & 75.42$_{(\pm0.08)}$ \\
    (0.8,0.5) & 46.10$_{(\pm31.91)}$      & 71.59$_{(\pm1.77)}$      & 77.39$_{(\pm0.09)}$ \\
    (0.5,0.5) & 35.77$_{(\pm34.76)}$      & 74.61$_{(\pm0.56)}$      & 77.00$_{(\pm0.29)}$ \\
    (0.5,0.2) & 73.10$_{(\pm1.72)}$       & 75.72$_{(\pm1.47)}$      & 77.11$_{(\pm0.08)}$ \\

    \bottomrule
    \end{tabular}%
     \caption{The testing accuracy (mean $\pm$ std \%) of ResNet164 on CIFAR100. $(N_a, N_b)$ is a pair of constant noise added to BN at the batch-normalized step as stated in Eqn.~\ref{eqn:Constant noise}. $(0.0, 0.0)$ means we do not add the noise.}
  \label{tab:Constant noise}%
\end{table}
For IEBN, we can rewrite Eqn.~\ref{eqn:Constant noise2} as
\begin{equation}
        \hat{X}_{bc} = \Delta \cdot (N_{a} \cdot \gamma \cdot \delta_{bc}) + (N_{b}\cdot \gamma \cdot \delta_{bc} + \beta),
        \label{eqn:Constant noise3}
\end{equation}
where $\delta_{bc}$ denotes the attention learned in IEBN. Compared to Eqn.~\ref{eqn:Constant noise2}, Eqn.~\ref{eqn:Constant noise3} with $\delta_{bc}$ from IEBN has successfully adjusted constant noise and even achieved better performance under partial noise configuration. If $\delta_{bc}$ only excites $\beta$, we can rewrite Eqn.~\ref{eqn:Constant noise3} as
\begin{equation}
        \hat{X}_{bc} = \Delta \underbrace{\cdot N_a \cdot \gamma }_{\gamma'} + \underbrace{N_b\cdot \gamma  + \beta\cdot \delta_{bc}}_{\beta'},
        \label{eqn:Constant noise4}
\end{equation}

where $\delta_{bc}$ can only adjust the noise in $\beta'$ instead of $\gamma'$. But if applied to $\gamma$, $\delta_{bc}$ can handle the noise of scale and bias simultaneously. It may be the reason why the result about only exciting $\beta$ is worse than the other in Table~\ref{tab:leftright}~(Right), but better than the original model with BN in Table~\ref{tab:experiment}.

\subsection{Mix-Datasets Attack} In this part, we consider interfering with $\mu^B_c$ and $\sigma_c^B$ by simultaneously training on the datasets with different distributions in one network. Unlike constant noise which is added to networks directly, this noise is implicit and is generated when BN computes the mean and variance of training data from different distribution. These datasets differ widely in their distribution and causes severe batch noise. Compared with the constant noise, this noise is not easy to eliminate by linear transformation of $\gamma$ and $\beta$.

In our experiments, we train ResNet164 on CIFAR100 but mix up with MINIST~\cite{lecun-mnisthandwrittendigit-2010} or FashionMINIST~\cite{xiao2017/online} in a batch and compare the performance of BN and IEBN. Table~\ref{tab:tri_attack} shows the test accuracy on CIFAR100. As $k$ increases, the batch noise becomes more severe for CIFAR100 since $\mu^B_c$ and $\sigma^B_c$ contains more information about MNIST or FashionMnist. In most cases, despite the severe noise like ``C+2$\times$'', the model with IEBN still performs better than the model with BN training merely on CIFAR100. On the other hand, the drop in accuracy of IEBN is smaller than that of BN, and IEBN alleviates the degradation of network generalization. These phenomena illustrate that the model with IEBN has a stronger ability to resist the batch noise even if under the influence of MINIST or FashionMINIST.

\begin{table}[htbp]
  \centering
    \begin{tabular}{|l|c|c|c|c|}
    \toprule
          & \multicolumn{2}{c|}{BN} & \multicolumn{2}{c|}{IEBN} \\
\cmidrule{2-5}    \multicolumn{1}{|c|}{Dateset} & test acc  & acc drop  & test acc  & acc drop  \\
    \midrule
    C & 74.29 & -0.00 & 77.09 & -0.00 \\
    C+2$\times$ M & 73.13 & -1.16 & 76.65 & -0.44 \\
    C+3$\times$ M & 71.54 & -2.75  & 76.03 & -1.06 \\
    C+2$\times$ F & 71.56 & -2.73 & 75.57 & -1.52 \\
    C+3$\times$ F & 71.27 & -3.02 & 74.26 & -2.83 \\
    \bottomrule
    \end{tabular}%
  \caption{Test accuracy (\%) on CIFAR100 with ResNet-164. ``C+$k\times$ M/F'' means we samples a batch consisted of 100 images from CIFAR100 (C) and $120\times k$ images from MNIST (M) or FashionMNIST (F) at each iteration during training. ``acc drop'' means the drop of accuracy compared with network trained merely on CIFAR100.}
  \label{tab:tri_attack}%
\end{table}%

\subsection{Ablation Study}
In this section, we conduct experiments to explore the effect of different configurations of IEBN.
All experiments are performed on CIFAR100 with ResNet164 using 2 GPUs.

\subsubsection{The Way of Generating $\delta_{bc}$.}
This part we study different ways to process the squeezed features to generate $\delta_{bc}$. As shown in Alg.~\ref{alg:IEBN}, IEBN squeezes the channel through global average pooling $\text{AVG}(\cdot)$ and processes the squeezed feature by linear transformation~(i.e.~AVG($X_{bc}$)$\times \hat{\gamma}_c$ + $\hat{\beta}_c$) for each channel, denoted as ``Linear''. We also consider another two methods to process the information. The first one is that we remove the additional trainable parameters $\hat{\gamma}_c$ and $\hat{\beta}_c$ for linear transformation in IEBN and directly apply the squeezed feature after sigmoid function to the channel, denoted as ``Identity''. The second one is that we use a fully connected layer stacking of a linear transformation, a ReLU layer, and a linear transformation to fuse the squeezed features of all channels $\{\text{AVG}(X_{bc})\}_{c=1}^C$, denotes as ``FC''. ``FC'' is similar to the configuration as SE module introduced in \cite{hu2018squeeze}. As shown in Table~\ref{tab:leftright}~(Left), ''FC'' operator provides more nonlinearity than ``Linear'' operator~(IEBN), but such nonlinearity may lead to overfitting and the ``Linear'' operator~(IEBN) simplifies the squeezed feature processing and has better generalization ability. Furthermore, the result of "Identity" indicates that it is not enough to simply and directly use instance information to enhance self-information without any trainable parameters. The operators with trainable parameters, such as ``Linear''~(IEBN) and ``FC'', are needed to process the instance information such that the adaptive and advantageous noise during training can be regulated to improve the performance.

\begin{table}
  \centering
  \small
    \begin{tabular}{|c|c|c|c|c|}
\cmidrule{1-2}\cmidrule{4-5}    Operator & Test Acc.     &       & Position & Test \\
\cmidrule{1-2}\cmidrule{4-5}   Linear  & \textbf{77.09($\pm0.15$)}     &       & $\gamma$ & \textbf{77.09($\pm0.15$)} \\
    Identity  & 67.53($\pm$2.49)     &       & $\beta$  & 75.03($\pm0.54$) \\
   FC     & 76.11($\pm0.28$)     &       & $\gamma$ and $\beta$ & 77.02($\pm0.08$) \\
\cmidrule{1-2}\cmidrule{4-5}    \end{tabular}%
   \caption{(\textbf{Left}.)~Testing accuracy (\%) with different ways to process the squeezed features. (\textbf{Right}.)~Testing accuracy (\%) with different positions that the $\delta_{bc}$ excites. $\gamma$ and $\beta$ are the parameters in reparameterization step of BN on CIFAR100.}
\label{tab:leftright}
\end{table}%

\subsubsection{Excitation Position.}
We study the influence of different positions that $\delta_{bc}$ excites. For self-attention mechanism like SENet~\cite{hu2018squeeze}, the rescaling coefficient usually excites both the trainable parameter $\gamma$ and $\beta$ of BN. In IEBN, the $\delta_{bc}$ is only applied to adjust the scaling parameter $\gamma$ in BN. To differentiate the influence of the excitation positions, Table~\ref{tab:leftright}~(Right) shows testing accuracy with different positions where the $\delta_{bc}$ excites. We show that the performance is unsatisfied when the $\delta_{bc}$ is merely exciting $\beta$. Moreover, there is a slight difference between exciting only $\gamma$ and exciting both $\gamma$ and $\beta$, and the former excitation position has better performance. From the point of view of adjusting noise, Eqn.~\ref{eqn:Constant noise3} and Eqn.~\ref{eqn:Constant noise4} can explain the result shown in Table~\ref{tab:leftright}~(Right).

\subsubsection{Initialization of $\hat{\gamma}_c$ and $\hat{\beta}_c$}
We study the initialization of parameters $\hat{\gamma}_c$ and $\hat{\beta}_c$ which are used to process the squeezed feature in IEBN. Specifically, we use constant 1, 0 and -1 for grid search to find the best pair of initialization for $\hat{\gamma}_c$ and $\hat{\beta}_c$. The initialization of the trainable parameters of IBEN $\hat{\gamma}_c$ and $\hat{\beta}_c$ have a significant impact on the model.
The best choice of $\hat{\gamma}_c$ and $\hat{\beta}_c$ are 0 and -1. The theoretical nature behind the best initialization configuration will be our future work.

\begin{table}[htbp]
  \centering
  \small
    \begin{tabular}{|c|c|c|c|}
    \toprule
    $\hat{\gamma}_c\setminus \hat{\beta}_c$   & 1     & 0     & -1 \\
    \midrule
    1     & 68.5  & 66.96 & 69.53 \\
    0     & 75.86 & 76.21 & \textbf{77.09} \\
    -1    & 74.64 & 74.73 & 75.31 \\
    \bottomrule
    \end{tabular}%
  \caption{Test accuracy (\%) with different constant initialization for trainable parameters scaling $\hat{\gamma}_c$ and shift $\hat{\beta}_c$ in IEBN.}
  \label{tab:Initialization}%
\end{table}%

\subsubsection{Activation Function.}
\begin{figure}
  \centering
  \includegraphics[height=1.5in, width=2.5in]{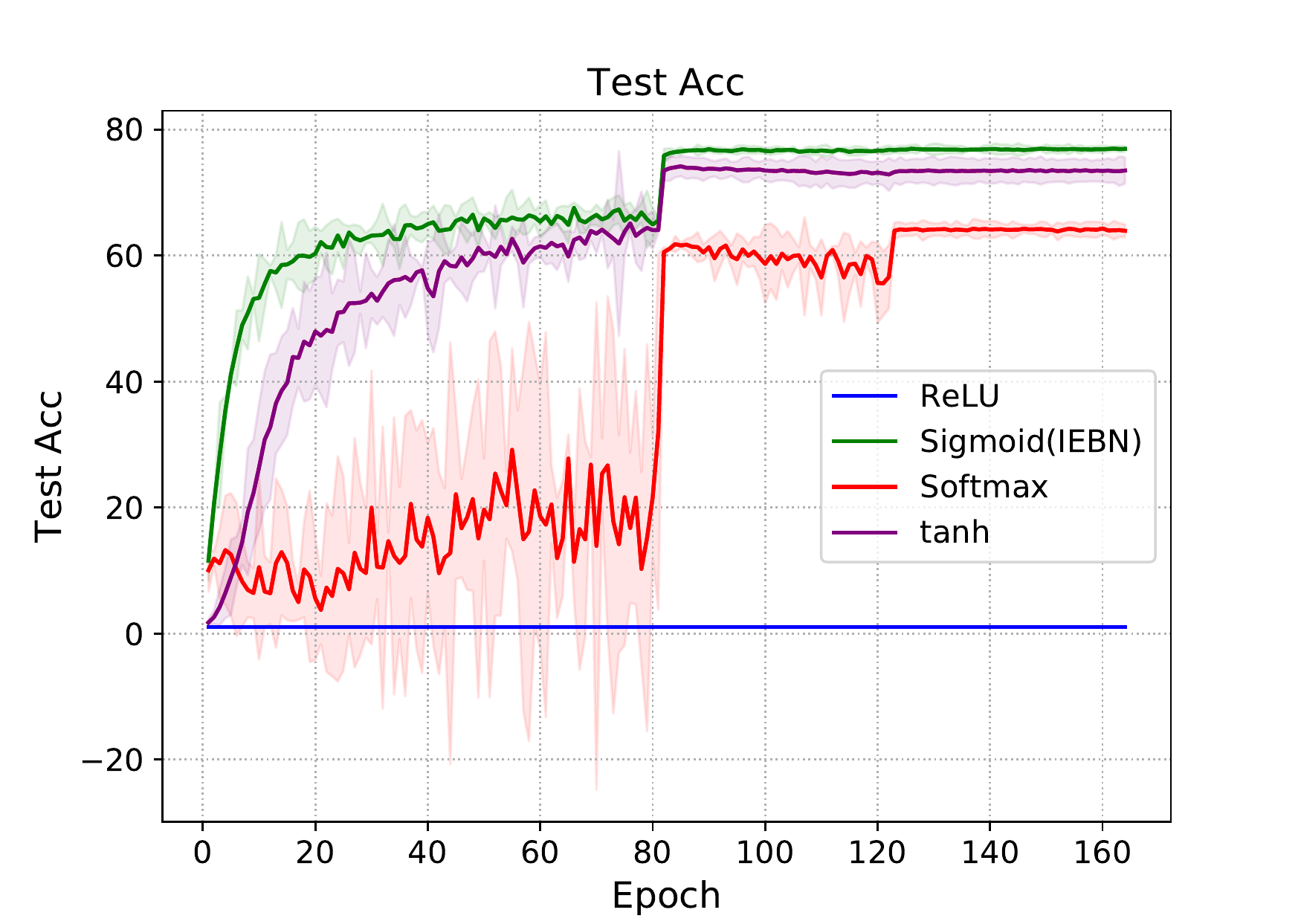}\\
  \caption{The training curve with different activation functions in IEBN.}\label{fig:activation_acc}
\end{figure}

 We consider four options for activation function: sigmoid, tanh, ReLU and Softmax in IEBN. The testing accuracy results are reported in Fig.~\ref{fig:activation_acc}. Note that, In these options, sigmoid may be the best choice, which is also used in many attention-based methods like SENet~\cite{hu2018squeeze} to generate attention maps as a gate mechanism. The testing accuracy of different choices of activation functions in Table~\ref{fig:activation_acc} shows that sigmoid helps IEBN as a gate to rescale channel features better. The similar ablation study in the SENet paper~\cite{hu2018squeeze} also shows the performance of different activation functions like: sigmoid, $>$ tanh $>$, and ReLU~(bigger is better), which coincides to our reported results.

\section{Conclusion}
In this paper, we introduce two kinds of noise brought by BN and offer a point of view that self-attention mechanism can regulate the batch noise adaptively. We propose a simple-yet-effective and attention-based BN called as Instance Enhancement Batch Normalization~(IEBN). We demonstrate empirically the effectiveness of IEBN on benchmark datasets with different network architectures and also provide ablation study to explore the effect of different configurations of IEBN.
\bibliographystyle{aaai}
\section{Acknowledgments}
	
 S. Liang and H. Yang gratefully acknowledge the support of National Supercomputing Center (NSCC) Singapore \cite{nscc} and High Performance Computing (HPC) of National University of Singapore for providing computational resources, and the support of NVIDIA Corporation with the donation of the Titan Xp GPU used for this research. Z. Huang thanks New Oriental AI Research Academy Beijing for GPU resources. H. Yang thanks the support of the start-up grant by the Department of Mathematics at the National University of Singapore, the Ministry of Education in Singapore for the grant MOE2018-T2-2-147.

\newpage
\section{Appendix}
\subsection{Implementation Detail}
The implementation detail is shown in Table~\ref{tab:detail1} and Table~\ref{tab:detail2}.
\begin{table*}[h]
        \small
        \centering
        \begin{tabular}{|c|c|c|c|c|}
            \toprule
            & ResNet164 & PreResNet164 & ResNext29-8x64 & Densenet100-12\\
            \midrule
            Batch size & 128   & 128   & 128 & 64\\
            Epoch & 180   & 164    & 300 & 300\\
            Optimizer & SGD(0.9) & SGD(0.9) & SGD(0.9) & SGD(0.9)\\
            depth & 164   & 164    & 29 & 100\\
            schedule & 81/122 & 81/122  & 150/225 & 150/225\\
            wd    & 1.00E-04 & 1.00E-04  & 5.00E-04 & 1.00E-04\\
            gamma & 0.1   & 0.1   & 0.1  & 0.1\\
            widen-factor & -     & -     & 4 &-\\
            cardinality & -     & -        & 8 & - \\
            lr    & 0.1   & 0.1    & 0.1 & 0.1\\
            \bottomrule
        \end{tabular}%
        \caption{Implementation detail for \textbf{CIFAR10/100} image classification. Normalization and standard data augmentation (random cropping and horizontal flipping) are applied to the training data.}
        \label{tab:detail1}%
        \vspace{-0.2cm}
    \end{table*}%

    \begin{table*}[h]
        \centering
        \begin{tabular}{|c|c|c|c|c|}
            \toprule
            & ResNet34 & ResNet50 & ResNet152 & ResNext50-32x4\\
            \midrule
            Batch size & 256   & 256   & 256   & 256 \\
            Epoch & 120   & 120   & 120   & 120 \\
            Optimizer & SGD(0.9) & SGD(0.9) & SGD(0.9) & SGD(0.9) \\
            depth & 34    & 50    & 152   & 50 \\
            schedule & 30/60/90 & 30/60/90 & 30/60/90 & 30/60/90 \\
            wd    & 1.00E-04 & 1.00E-04 & 1.00E-04 & 1.00E-04 \\
            gamma & 0.1   & 0.1   & 0.1   & 0.1 \\
            lr    & 0.1   & 0.1   & 0.1   & 0.1 \\
            \bottomrule
        \end{tabular}%
        \caption{Implementation detail for \textbf{ImageNet 2012} image classification. Normalization and standard data augmentation (random cropping and horizontal flipping) are applied to the training data. The random cropping of size 224 by 224 is used in these experiments.}
        \label{tab:detail2}%
        \vspace{-0.18cm}
    \end{table*}%
\subsection{Other Style Transfer Loss}
\begin{figure*}[h]
    \centering
    \includegraphics[height=5in, width=6in]{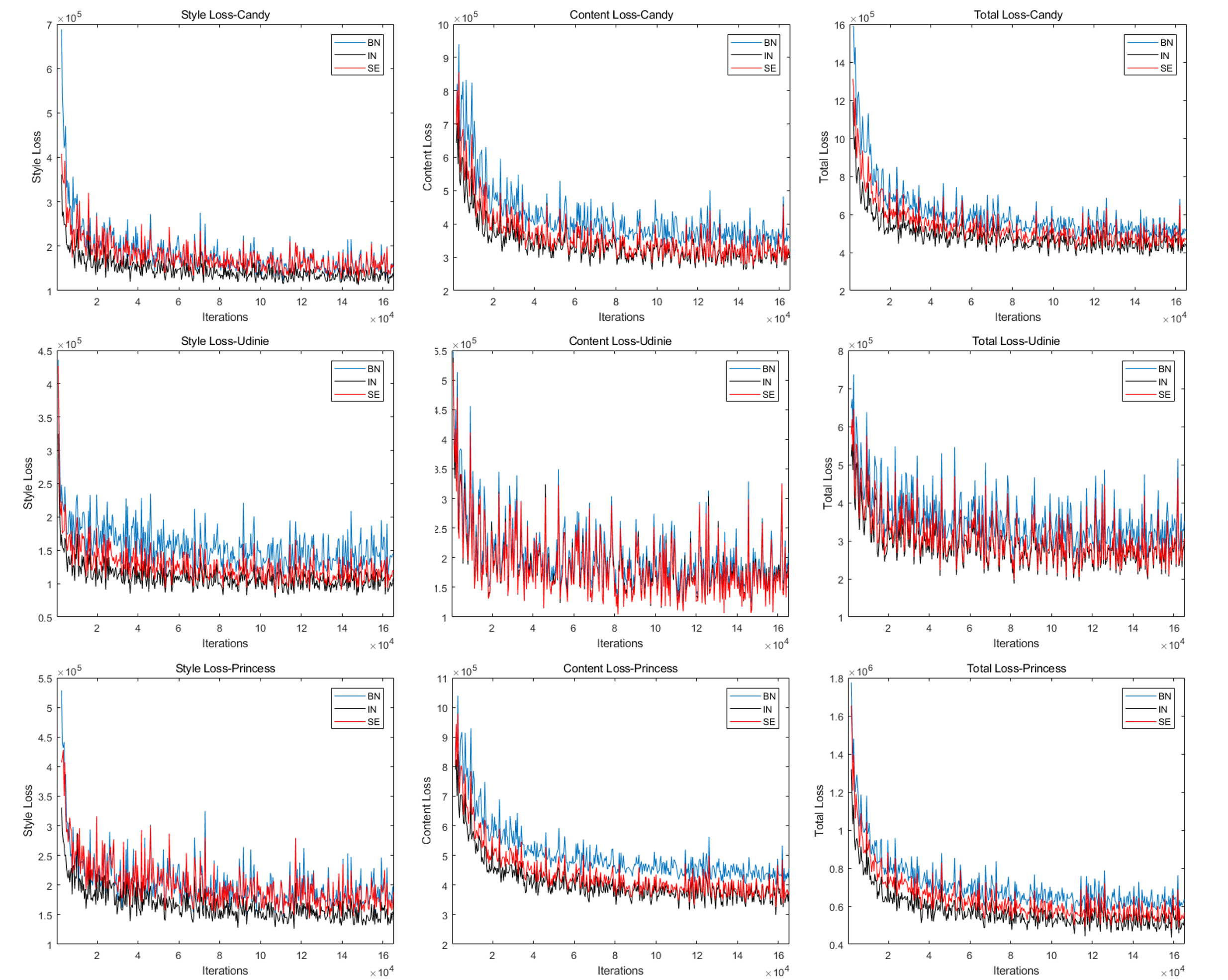}
    \caption{Training curves of style transfer networks with different styles and different normalization methods. Specially, ``SE'' means the transformation network with BN and SE module.}
    \label{fig:others}
\end{figure*}

The style transfer loss of different styles can be found in Fig.~\ref{fig:others}.
\end{document}